\documentclass[10pt,twocolumn,letterpaper]{article}

\usepackage[pagenumbers]{cvpr} 

\usepackage{textcomp}
\usepackage{multirow}
 \usepackage{multicol}
\usepackage{stfloats}
\usepackage{url}
\usepackage{verbatim}
\usepackage{graphicx}
\usepackage{amsmath}
\usepackage{amssymb}
\usepackage{booktabs}
\usepackage{color, colortbl}
\usepackage{verbatim}
\usepackage{cite}
\usepackage{booktabs}
\usepackage{tabularx}
\usepackage[dvipsnames]{xcolor}

\definecolor{cvprblue}{rgb}{0.21,0.49,0.74}
\usepackage[pagebackref,breaklinks,colorlinks]{hyperref}
\usepackage{xcolor}

\def\paperID{6784} 
\def\confName{CVPR}
\def\confYear{2024}

\newcommand{\tg}[1]{{\color{magenta} #1}}

\title{SGV3D: Towards Scenario Generalization for Vision-based Roadside\\ 3D Object Detection}
\author{
Lei Yang\textsuperscript{1}, Xinyu Zhang\textsuperscript{1}\footnotemark[1], Jun Li\textsuperscript{1}, Li Wang\textsuperscript{2},  Chuang Zhang\textsuperscript{1}, Li Ju\textsuperscript{3}, Zhiwei Li\textsuperscript{4}, Yang Shen\textsuperscript{1} \\
\textsuperscript{1} School of Vehicle and Mobility,  Tsinghua University;
\textsuperscript{2} Beijing Institute of Technology \\
\textsuperscript{3} National University of Singapore;
\textsuperscript{4} Beijing University of Chemical Technology \\
    \begin{normalsize}${\tt \{yanglei20@mails., lijun1958@, xyzhang@, zhch20@mails., shenyang@mail.\}tsinghua.edu.cn}$\end{normalsize} \\
    \begin{normalsize}${\tt wangli\_bit@bit.edu.cn; e1143667@u.nus.edu; lizw@buct.edu.cn}$\end{normalsize}}

\begin{document}
\maketitle

\renewcommand{\thefootnote}{\fnsymbol{footnote}}
\footnotetext[1]{Corresponding Author.}
\renewcommand{\thefootnote}{\arabic{footnote}}

\begin{abstract}
Roadside perception can greatly increase the safety of autonomous vehicles by extending their perception ability beyond the visual range and addressing blind spots. However, current state-of-the-art vision-based roadside detection methods possess high accuracy on labeled scenes but have inferior performance on new scenes. This is because roadside cameras remain stationary after installation and can only collect data from a single scene, resulting in the algorithm overfitting these roadside backgrounds and camera poses. To address this issue, we propose an innovative \textbf{S}cenario \textbf{G}eneralization Framework for \textbf{V}ision-based Roadside \textbf{3D} Object Detection, dubbed \textbf{SGV3D}. Specifically, we employ a Background-suppressed Module (BSM) to mitigate background overfitting in vision-centric pipelines by attenuating background features during the 2D to bird's-eye-view projection. Furthermore, by introducing the Semi-supervised Data Generation Pipeline (SSDG) using unlabeled images from new scenes, diverse instance foregrounds with varying camera poses are generated, addressing the risk of overfitting specific camera poses. We evaluate our method on two large-scale roadside benchmarks. Our method surpasses all previous methods by a significant margin in new scenes, including +42.57\% for vehicle, +5.87\% for pedestrian, and +14.89\% for cyclist compared to BEVHeight on the DAIR-V2X-I heterologous benchmark. On the larger-scale Rope3D heterologous benchmark, we achieve 
notable gains of 14.48\% for car and 12.41\% for large vehicle. The code will be available at \href{https://github.com/yanglei18/SGV3D}{SGV3D}.

\vspace{-0.58cm}
\end{abstract}
\section{Introduction}
\label{sec:intro}
\begin{figure}[h]
	\centering
	\includegraphics[width=8.3cm]{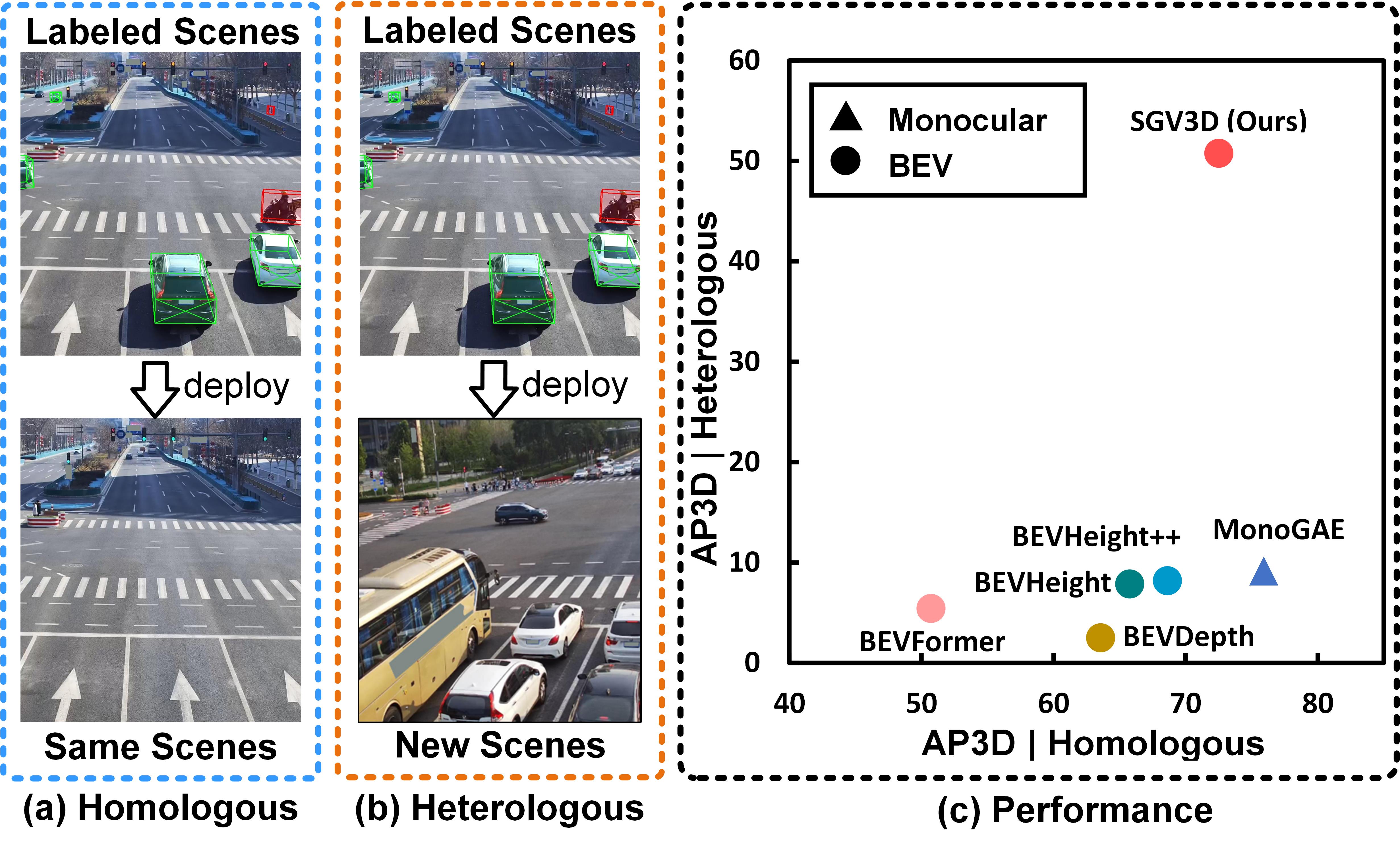}
	\caption{(a) The homogeneous setting entails utilizing images from the same roadside scenes for both training and testing. (b) The heterogeneous setting involves training with images from labeled scenes but testing in entirely new and diverse scenes. (c) Presently, vision-based roadside 3D object detection methods demonstrate high performance in the homogeneous setting but experience a significant accuracy decline in the heterogeneous setting. Our SGV3D surpasses all state-of-the-art methods by a substantial margin in the heterogeneous setting, highlighting robust scenario generalization.}
\label{fig:teaser}
\vspace{-0.45cm}
\end{figure}
\begin{figure*}[t!]
	\centering
	\includegraphics[width=17.8cm]{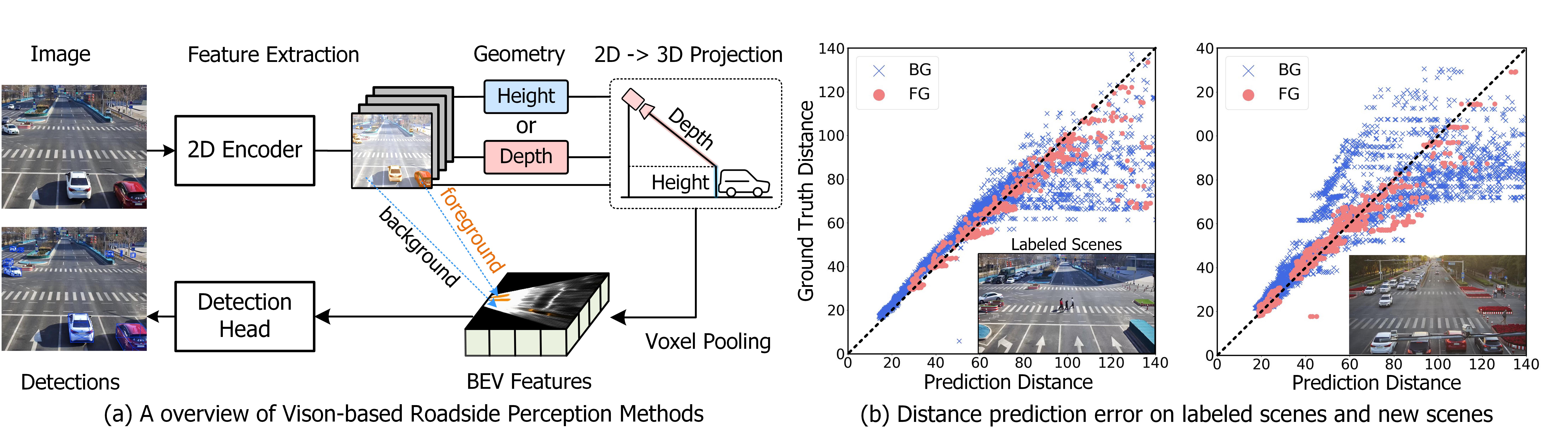}
	\caption{\textbf{Empirical analysis on the poor scenario generalization of existing methods.} (a) We present an overview of previous vision-centric roadside 3D object detectors. In particular, the background regions constitute the majority of BEV features. (b) We plot the pixel-level scatter diagram revealing the distance correlation between ground truth and predicted distance based on the BEVHeight. Here, we convert height to distance for a more intuitive comparison. The distance errors for the background regions marked as `BG' in the new scenes are significantly larger than the errors in the labeled scenes. The foreground distance errors marked as `FG' in new scenes are more pronounced compared to those in labeled scenes as well.}
 \vspace{-0.25cm}
\label{fig:intro}
\end{figure*}
3D object detection \cite{yang2023lite,wang2023multi,song2023graphalign++,song2024robustness,song2024robofusion,song2024voxelnextfusion,song2023vp,zhang2023urformer,9726893} in autonomous driving has made significant progress in recent years. Nevertheless, it still faces significant challenges, including blind spot occlusion and limited long-range perception capability due to the restricted installation height of on-board cameras. Recently, with the rapid development of intelligent infrastructure, it has become possible to use roadside cameras for 3D object detection. Taking advantage of the increased height compared to on-board sensors,  roadside cameras can realize a broader field of view and are less susceptible to vehicle occlusion. Assisted by roadside perception \cite{9764653, yang2023bevheight, yang2023bevheight++, shi2023cobev, fan2023calibration, yang2023monogae} autonomous vehicles can achieve a global perspective that extends well beyond the current horizon and eliminates blind spots through cooperative techniques~\cite{ yu2023flow, song2023spatial, ruan2023learning,fan2023quest,10265751,9228884,10106610, li2023among,10122468,li2023multi,10398509}, thereby significantly improving safety.

To facilitate roadside perception research, some large-scale benchmark \cite{yu2022dair,ye2022rope3d,yu2023v2x,li2022v2x-sim} with diverse camera data have been introduced, along with evaluations of baseline methods. A number of vision-based roadside 3D object detectors \cite{yang2023bevheight,yang2023bevheight++,yang2023monogae,shi2023cobev,jia2023monouni} have also emerged in recent years. However, current state-of-the-art vision-based roadside detection methods exhibit poor transfer ability in new scenes. As shown in Fig.~\ref{fig:teaser}, the performance of existing methods trained on labeled scenes is accurate in the same scenes but shows lower performance in new scenes. 
This is because roadside cameras remain fixed after installation, and collect images solely from a single scene, causing the algorithm to overfit these specific roadside backgrounds and camera poses. Obviously, annotating images from all new roadside scenes is both costly and impractical, thus hindering the large-scale deployment of roadside perception systems.

Current vision-centric methods, as described in \cite{yang2023bevheight}, can be shown in Fig.~\ref{fig:intro} (a), the original image is first converted into 2D feature maps with a 2D encoder. These feature maps are then used to predict the per-pixel depth or height of the feature map. Each pixel feature can be lifted into 3D space and zipped into the BEV feature space using voxel pooling techniques. The resulting BEV features are then fed into the detection head to generate the final predictions. 
We conducted a detailed analysis of the pixel-level distance prediction errors. As shown in Fig.~\ref{fig:intro} (b), the scatter plots representing both foreground and background within the image of the labeled scene closely follow the diagonal line. In contrast, for the images collected from new scenes, the scatterplot of the background shows a divergent trend that deviates significantly from the diagonal line. Owing to these notable distance errors in the background regions, which constitute the majority of the BEV features, accomplishing a precise transformation of the background features from the image view to the bird's eye view poses a substantial challenge. Meanwhile, the foreground distance errors in new scenes are more pronounced than those in labeled scenes, This discrepancy is attributed to the varied camera poses across different scenes. All these factors contribute to the pool transferability to new scenes of current vision-centric methods. 

In this paper, we propose \textbf{SGV3D}, a novel \textbf{S}cenario \textbf{G}eneralization Framework for \textbf{V}ision-based Roadside \textbf{3D} Object Detection. Our strategy, termed ``background suppression and foreground enrichment," addresses the challenges of overfitting to background and specific camera poses within labeled scenes. For background suppression, we introduce a specialized background-suppressed BEV detector built upon BEVHeight \cite{yang2023bevheight}, predicting height for image-to-BEV view transformation. The Background-Suppressed Module (BSM) within this detector effectively mitigates the influence of background features during the transition from 2D to bird's-eye-view. This is achieved through a semantic segmentation branch that enhances image features while suppressing background elements. For foreground enrichment, we implement a semi-supervised data generation pipeline (SSDG) utilizing unlabeled images from new scenes. This process involves combining instances with high-quality pseudo-labels and empty background images, yielding well-labeled images featuring diverse foreground instances under various camera poses. The synthetic dataset, in conjunction with annotated data, is then utilized to train the background-suppressed detector described above. This streamlined pipeline, reliant solely on unlabeled data from new scenes, avoids the need for resource-intensive manual annotation, thereby facilitating the widespread deployment of roadside perception systems.

We conducted extensive experiments on the DAIR-V2X-I~\cite{yu2022dair} and Rope3D~\cite{ye2022rope3d} datasets. For the DAIR-V2X-I benchmark, our SGV3D demonstrates state-of-the-art performance under homogeneous conditions. When subjected to a heterogeneous setting, existing state-of-the-art methods exhibit a significant decrease in accuracy compared to their performance under homogeneous conditions. For example, the BEVDepth \cite{li2023bevdepth} experiences a reduction from 63.58\% to 2.48\%, BEVHeight \cite{yang2023bevheight} from 65.77\% to 7.86\%. This decline highlights the inadequacy of these methods in handling scene adaptation. In contrast, our SGV3D maintains accuracy, surpassing leading methods by +42.57\%, +5.87\%, and +14.89\% for vehicle, pedestrian, and cyclist. Under the larger-scale Rope3D heterologous benchmark, we achieve substantial gains, with increases of +14.48\% and +12.41\% for the car and big vehicle, respectively. All these experiments certify the strong scenario generalization capability of our method.
\section{Related Works}
\label{sec:relatedworks}

\noindent \textbf{Roadside Perception.} Roadside perception aims to enhance the safety of autonomous vehicles by extending their perceptual capabilities beyond the visual field and addressing blind spots. Pioneers have recently presented roadside datasets \cite{yu2022dair,ye2022rope3d,yu2023v2x,li2022v2x-sim} to facilitate 3D perception tasks in these scenarios, coinciding with the simultaneous emergence of numerous related methods \cite{yang2023bevheight, yang2023bevheight++, shi2023cobev, fan2023calibration, yang2023monogae, jia2023monouni}. BEVHeight \cite{yang2023bevheight} initially focuses on roadside detection and proposes a novel approach to predict the height distribution of the scene rather than the depth distribution, resulting in improved performance.  CBR \cite{fan2023calibration} uses multi-layer perceptrons (MLPs) to map image features to BEV space, bypassing extrinsic calibration, but with limited accuracy. CoBEV \cite{shi2023cobev} generates robust BEV representations by seamlessly incorporating complementary geometry-centric depth and semantic-centric height cues. Despite promising results in homogeneous settings, these methods fail to maintain high-precision metrics in heterogeneous configurations, particularly with respect to the scenario generalization of roadside perception, which remains unexplored.


\noindent\textbf{Scenario Generalization.} Scenario generalization for vision-based roadside 3D object detection can be conceptualized as a form of domain adaptation, wherein an object detector initially trained on a fully supervised source domain undergoes adjustment for a target domain. This study designates labeled scenes as the source domain, whereas new scenes are identified as the target domain. The predominant focus of research in this domain lies in source-free adaptation and semi-supervised adaptation. Source-free adaptation \cite{vs2023instance, liu2023periodically, lin2023run, hegde2023source} is tailored for scenarios in which source domain data is inaccessible during the adaptation process. Semi-supervised approaches \cite{wang2023ssda3d, yu2023semi, jain2023marrs, nie2023adapting, xie2023adapt, xu2023semi,zhang2023semi} embrace a teacher-student framework. In this paradigm, a teacher detector, trained on the source domain, generates pseudo-labels for target domain images. A student model is subsequently trained using these labels to augment its performance on the target domain. In this paper, we enhance the scenario generalization from two perspectives: ``background suppression and enriching foreground." The background suppression through a Background-Suppressed Module (BMS) aligns with source-free adaptation, while foreground enrichment by semi-supervised data generation pipeline (SSDG) corresponds to semi-supervised adaptation.
\section{Method}

\subsection{Problem Definition}
In this work, we focus on implementing a vision-based roadside 3D object detector with great scenario generalization ability, which can be defined as follows:
\begin{equation}
    \hat{B} = F_{Det}\left(X, E, I \vert \omega \right),
\label{con:eq1}
\end{equation}
where $F_{Det}\left(.\right)$ is the detector model, $\omega$ is the learned weights of the detector. $X\in \mathbb{R}^{H \times W \times3}$ is the color image collected by the roadside cameras. 
$H$ and $W$ represent the height and width of the image $X$. $E\in \mathbb{R}^{3 \times 4}$ and $I\in \mathbb{R}^{3 \times 3}$ represent the extrinsic and intrinsic parameters of the roadside cameras, respectively. $\hat{B}$ implies the predicted 3D bounding boxes of objects on the image in the ego coordinate system, which can be defined as follows:

\begin{equation}
    \hat{B} = \left\{\hat{b}^{1}, \hat{b}^{2}, ... \hat{b}^{n}  \right\},
\label{con:eq2}
\end{equation}
where $n$ is the number of predictions for image $X$. Each prediction $\hat{b}$ can be represented as:

\begin{equation}
    \hat{b} = \left(x, y, z, h, w, l, \theta, conf \right),
\label{con:eq3}
\end{equation}
where $\left(x, y, z\right)$ is the location of each 3D bounding box. $\left(l, w, h\right)$ denotes the cuboid's length, width, and height respectively. $\theta$ represents the yaw angle of each instance with respect to one specific axis. $conf$ is the confidence score.

To ensure great performance in the new roadside scenarios, we follow the semi-supervised setting, This involves utilizing both labeled data $\left\{X_i^l,E_i^l,I_i^l,B_i^l\right\}_{i=1}^{N_l}$ from the existing scenes and unlabeled data $\left\{X_i^u,E_i^u,I_i^u\right\}_{i=1}^{N_u}$ that collected from new roadside scenes. Here $N_l$ and $N_u$ are the numbers of labeled and unlabeled data, respectively. It's noteworthy that the unlabeled data mentioned above does not include the validation or testing set.

\begin{figure*}[t!]
	\centering
	\includegraphics[width=17.3cm]{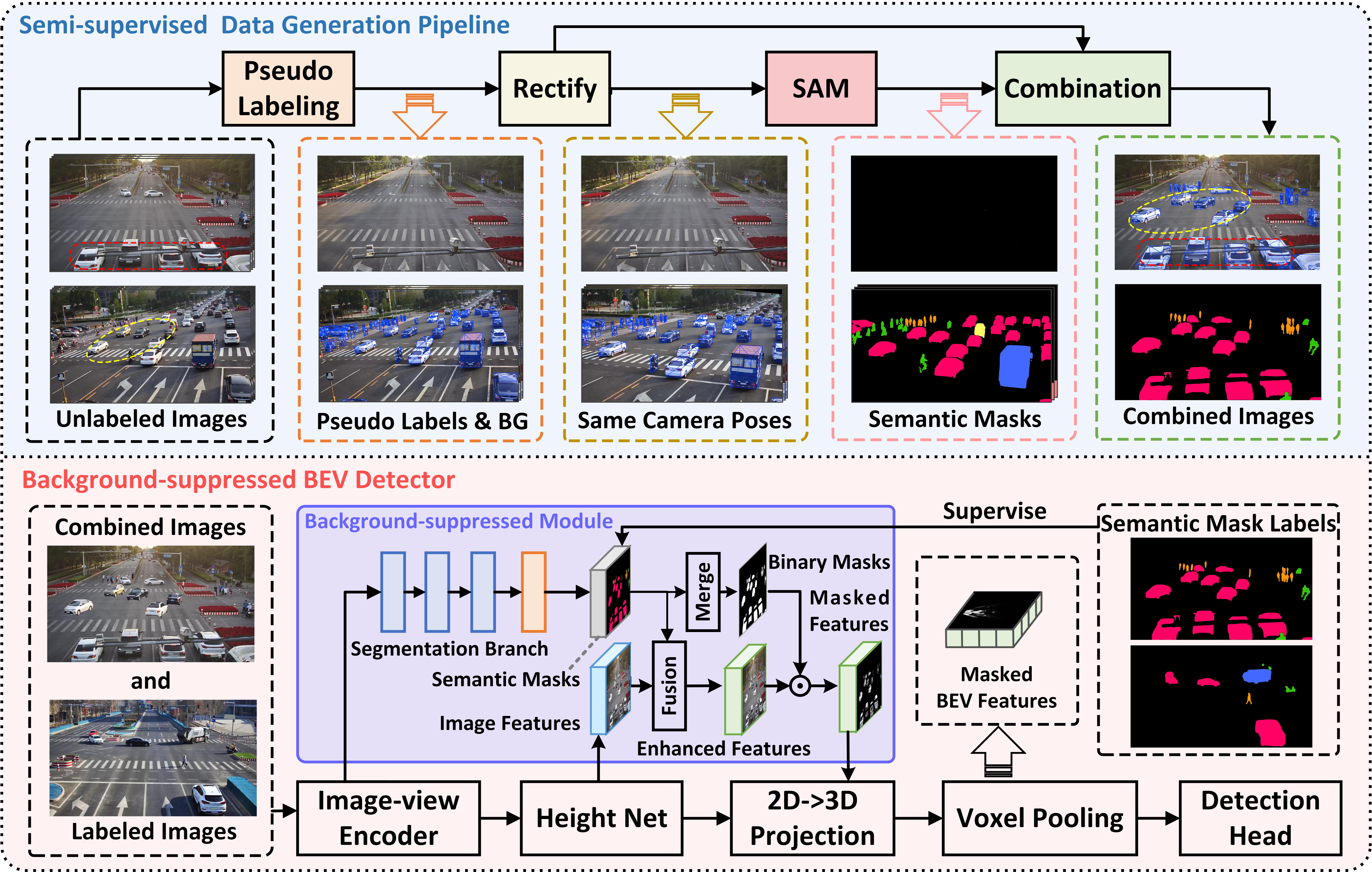}
	\caption{\textbf{The overall framework of SGV3D.} 
 the Background-suppressed BEV Detector with Background-suppressed Module (BMS) derived from the BEVHeight \cite{yang2023bevheight}. The Background-suppressed Module (BMS) reduces algorithm overfitting to the backgrounds of labeled scenes by suppressing background features with a semantic segmentation branch. Meanwhile, the Semi-supervised Data Generation Pipeline (SSDG) generates diverse, well-labeled images under different camera poses for the training stage, minimizing the risk of detector overfitting to specific camera settings, including intrinsic and extrinsic parameters. Our proposed framework performs multiple rounds of self-training strategy as STAC \cite{sohn2020simple}.}
 \label{fig:framework}
\end{figure*}

\subsection{SGV3D Architecture}
The illustration of our SGV3D is shown in Fig. \ref{fig:framework}, primarily consisting of two components: the Background-suppressed BEV Detector with Background-suppressed Module (BMS) and the Semi-supervised Data Generation Pipeline (SSDG). SSDG is responsible to generate diverse, well-labeled images under varying camera poses for Background-suppressed BEV Detector in the training process. We, at a high level, follow a multiple rounds of self-training scheme as STAC \cite{sohn2020simple}, where the initial teacher detector is trained on labeled data, followed by the SSDG process being applied to unlabeled data. A student detector is then trained on both manually labeled images and the images generated by SSDG. The resulting student detector subsequently serves as the teacher detector for SSDG in the next round of self-training.

The Background-suppressed BEV Detector adheres to the fundamental BEVHeight \cite{yang2023bevheight} pipeline. It incorporates a Background-suppressed Module (BMS) before the 2D to bird’s-eye-view projection module. The BMS effectively masks background region features, ensuring that the resulting BEV Features exclusively capture foreground region features. This mitigates the risk of overfitting to stationary backgrounds, enhancing the detector's adaptability to new scenes. Specifically, the Background-suppressed BEV Detector consists of six main modules and the BMS is a module we newly proposed. The image-view encoder, comprising a 2D backbone and an FPN module, is designed to extract 2D high-dimensional image features denoted as $F^{2d} \in \mathbb{R}^{C_F \times \frac{H}{16} \times \frac{W}{16}}$, given an image $X \in \mathbb{R}^{H \times W \times 3}$ in a roadside view. Here, $C_F$ represents the channels of image features. The HeightNet \cite{yang2023bevheight} is tasked with predicting a bins-like distribution of height from the ground $H^{pred} \in \mathbb{R}^{C_H \times \frac{H}{16} \times \frac{W}{16}}$ and context features $F^{context} \in \mathbb{R}^{C_c \times \frac{H}{16} \times \frac{W}{16}}$ based on the image features $F^{2d}$, where $C_H$ denotes the number of height bins, and $C_c$ represents the channels of the context features. The background-suppressed module predicts foregrounds semantic segmentation masks $M^{seg} \in \mathbb{R}^{C_s \times \frac{H}{16} \times \frac{W}{16}}$ based on the image features $F^{2d}$, where $C_s$ is the number of semantic classes. These semantic masks are then employed to generate semantic enhanced features $F^{enhc} \in \mathbb{R}^{C_{e} \times \frac{H}{16} \times \frac{W}{16}}$ and masked enhanced features $F_{mask}^{enhc} \in \mathbb{R}^{C_{e} \times \frac{H}{16} \times \frac{W}{16}}$, where $C_{e}$ is the channels of enhanced features. The height-based $2D\rightarrow 3D$ projector \cite{yang2023bevheight} transforms the masked enhanced features $F_{mask}^{enhc}$ into 3D wedge-shaped features $F^{wedge} \in \mathbb{R}^{X \times Y \times Z \times C_{e}}$ based on the predicted bins-like height distribution $H^{pred}$. Voxel Pooling \cite{reading2021categorical} converts the 3D wedge-shaped features into masked BEV features $F_{mask}^{bev}$ along the height direction. The 3D detection head initially encodes the BEV features with convolution layers and subsequently predicts the 3D bounding box, including location $(x, y, z)$, dimensions $(l, w, h)$, and orientation $\theta$.

\begin{figure}[t!]
	\centering
	\includegraphics[width=8.3cm]{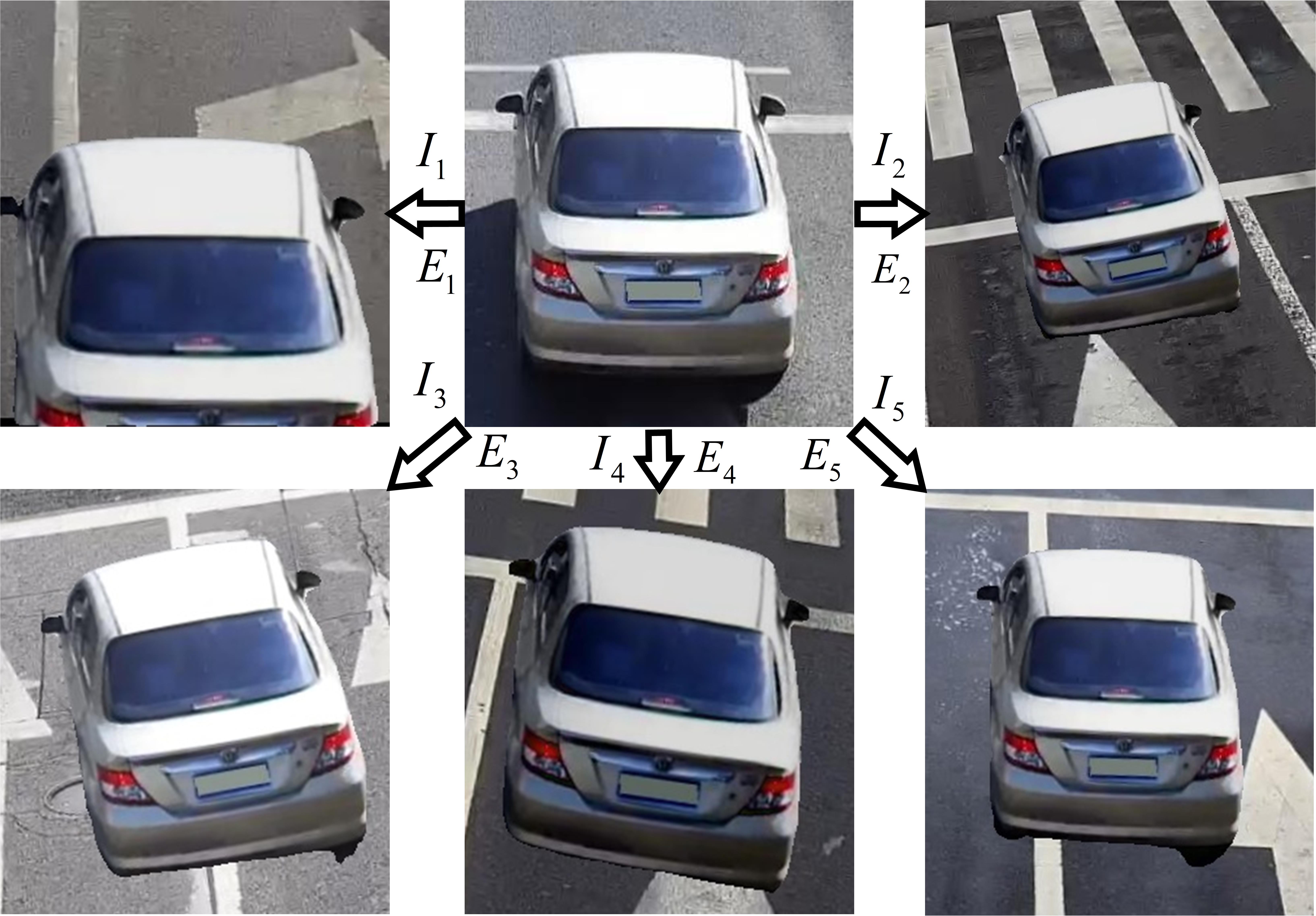}
	\caption{\textbf{Visualization of diverse foreground instances.} The same car, when captured by cameras with different intrinsic and extrinsic parameters, exhibits markedly different shapes and sizes.}
 \vspace{-0.01cm}
\label{fig:ssl_instance}
\end{figure}

The Semi-supervised Data Generation Pipeline relies on labeled data sourced from existing scenes and unlabeled data collected from new roadside scenes. By merging instances with high-quality pseudo-labels and empty background images, this pipeline produces well-labeled images featuring diverse foreground instances under various camera poses, as illustrated in Fig. \ref{fig:ssl_instance}. The synthetic dataset, coupled with pre-existing annotated data, is employed to train the Background-suppressed BEV Detector. This approach effectively mitigates the risk of detector overfitting to specific camera settings, such as intrinsic and extrinsic parameters, in labeled scenes. Consequently, it enhances the algorithm's generalization ability in new scenes. We employ a multi-round training paradigm, wherein the model from the preceding round is utilized for the pseudo-labeling process, and the generated data is subsequently integrated into the training process of this round.

\subsection{Background-suppressed Module}
By attenuating background features before the 2D to bird’s-eye-view projection, the Background-suppressed Module (BMS) aims to mitigate the risk of the algorithm overfitting to the backgrounds of labeled scenes.  It consists of three primary processes: multi-class semantic segmentation, feature fusion, and background filtering.

A segmentation branch is employed to predict multi-class semantic masks $M^{seg}$ for instance foreground.
This branch incorporates three residual blocks to enhance representation power and utilizes a deformable convolution layer to predict per-pixel multi-class semantic masks, as follows:
\begin{equation}
     M^{seg} = F_{Seg} \left(F^{2d}\right),
\label{con:eq4}
\end{equation}
where $F_{Seg}$ represents the segmentation branch.

Feature fusion combines context features $ F^{context}$ from HeightNet and the multi-class semantic masks $M^{seg}$ to obtain the semantic enhanced features $F^{enhc}$, which can formulated as follows:
\begin{eqnarray}    
F^{enhc} &=& F_{Fusion} (M^{seg}, F^{context})   \nonumber    \\
~& =& F_{Attn} (Conv([M^{seg}, F^{context}])),
\label{con:eq5}
\end{eqnarray}
where $[.,.]$ denotes the concatenation operation along the channel dimension, $Conv$ implies a $3 \times 3$ convolution layer to reduce the channel dimension into $C^{enhc}$, $F_{Attn}(F)$ is a channel attention layer to select important fused features and can be formulated as:
\begin{equation}
F_{Attn}(F) = \sigma(\Phi F_{Avg}(F)) \cdot F,
\label{con:eq6}
\end{equation}
where $\Phi$ is the linear transform matrix, $F_{Avg}$ denotes global average pooling, and $\sigma$ denotes the sigmoid function.

The masked features denoted as $F_{mask}^{enhc}$, are derived by filtering out background regions from the semantically enhanced features $F^{enhc}$, mathematically expressed as: 
\begin{equation}
    F_{mask}^{enhc} = M^{bin} \cdot F^{enhc},
\label{con:eq7}
\end{equation}
where $M^{bin}$ represents the binary mask of the instance foreground regions, and it is formulated as:
\begin{equation}
M^{bin} = \bigcup_{i=1}^{C_s -1} \left( M_i^{seg} > T_{fg}\right)
\label{con:eq8}
\end{equation}
where $T_{fg}$ is the threshold to identify foreground regions.

During the training process, the ground truth for the segmentation branch comes from the Segment Anything model (SAM), which uses 2D boxes as the prompt.

\subsection{Semi-supervised Data Generation Pipeline}
The Semi-supervised Data Generation Pipeline (SSDG) is designed to alleviate the risk of detector overfitting to specific camera settings, including intrinsic and extrinsic parameters associated with labeled scenes. The pipeline comprises four primary processes: Pseudo Labeling, Rectify, SAM, and Combination.

\noindent \textbf{Pseudo Labeling.} 
There are two objectives in this process. On the one hand, the pseudo-labels for unlabeled images are generated, where pseudo-labels are defined as predictions with confidence higher than the threshold $T_{conf}$, represented as follows:
\begin{eqnarray}
    \hat{B}^{u}&=&F_{Det}\left(X^u,E^u,I^u \vert \omega \right) \nonumber \\
              ~&=&\left\{\hat{b}^k |conf_k>T_{conf}\right\}_{k=1}^{n^p},
\label{con:eq9}
\end{eqnarray}
where $F_{Det}$ represents the detector model trained in the previous round. $T_{conf}$ implies the confidence threshold to select high-quality pseudo labels, $n^p$ is the number of pseudo labels for image $X^u$.

On the other hand, we follow \cite{8690428} to generate the empty background data $\left\{X_{bg}^{u}, E_{bg}^{u}, I_{bg}^{u} \right\}$, which will be used in the following combination process.

\begin{figure}[h!t]
	\centering
	\includegraphics[width=8.5cm]{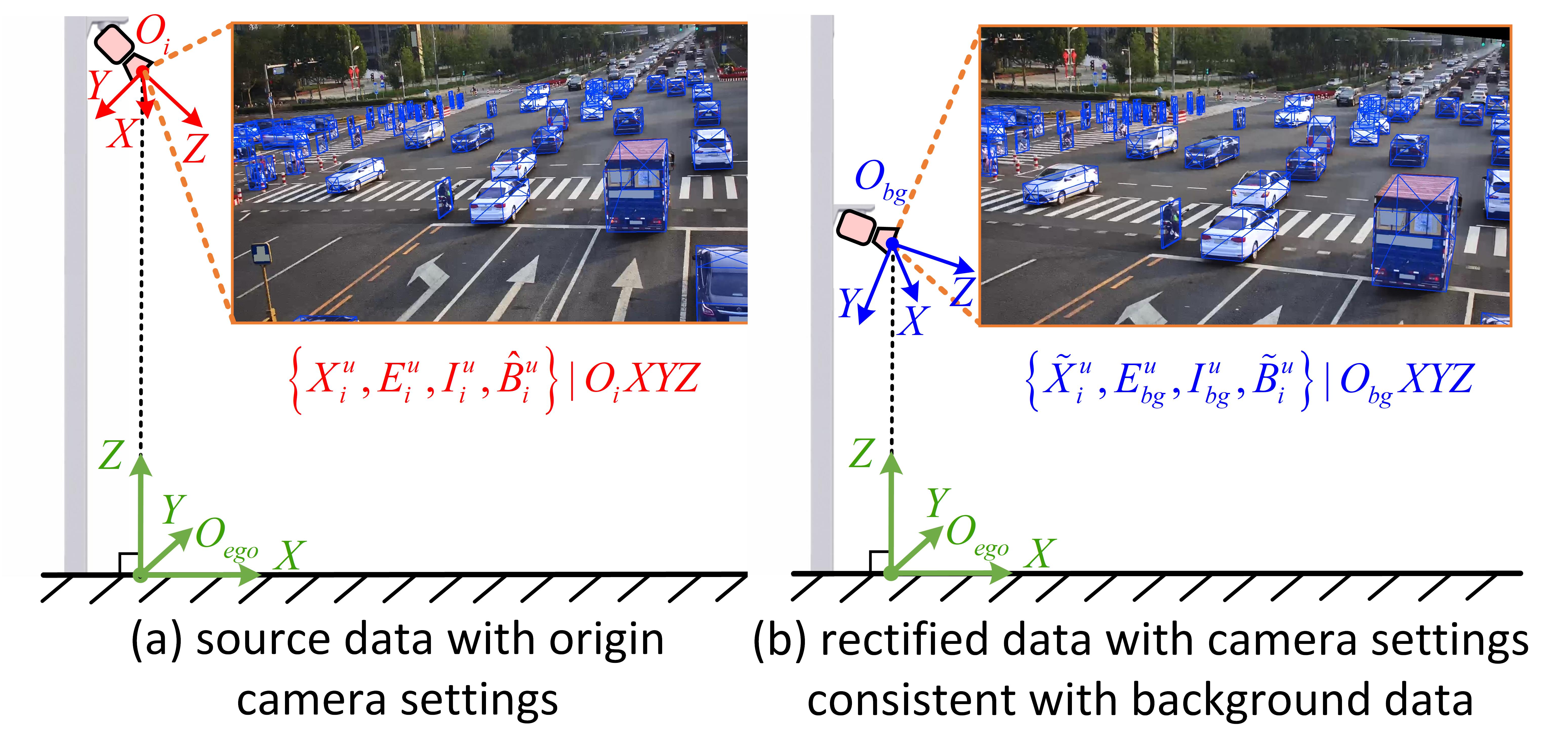}
	\caption{\textbf{Visualization of the source data and its rectified data.} {\color{red}$O_iXYZ$} implies the origin camera coordinate system. {\color{blue}$O_{bg}XYZ$} represents the rectified camera coordinate system, which is consistent with that of background data. }
\label{fig:problem_of_rectify}
\vspace{-0.15cm}
\end{figure}
\begin{figure}[h!t]
	\centering
	\includegraphics[width=8.5cm]{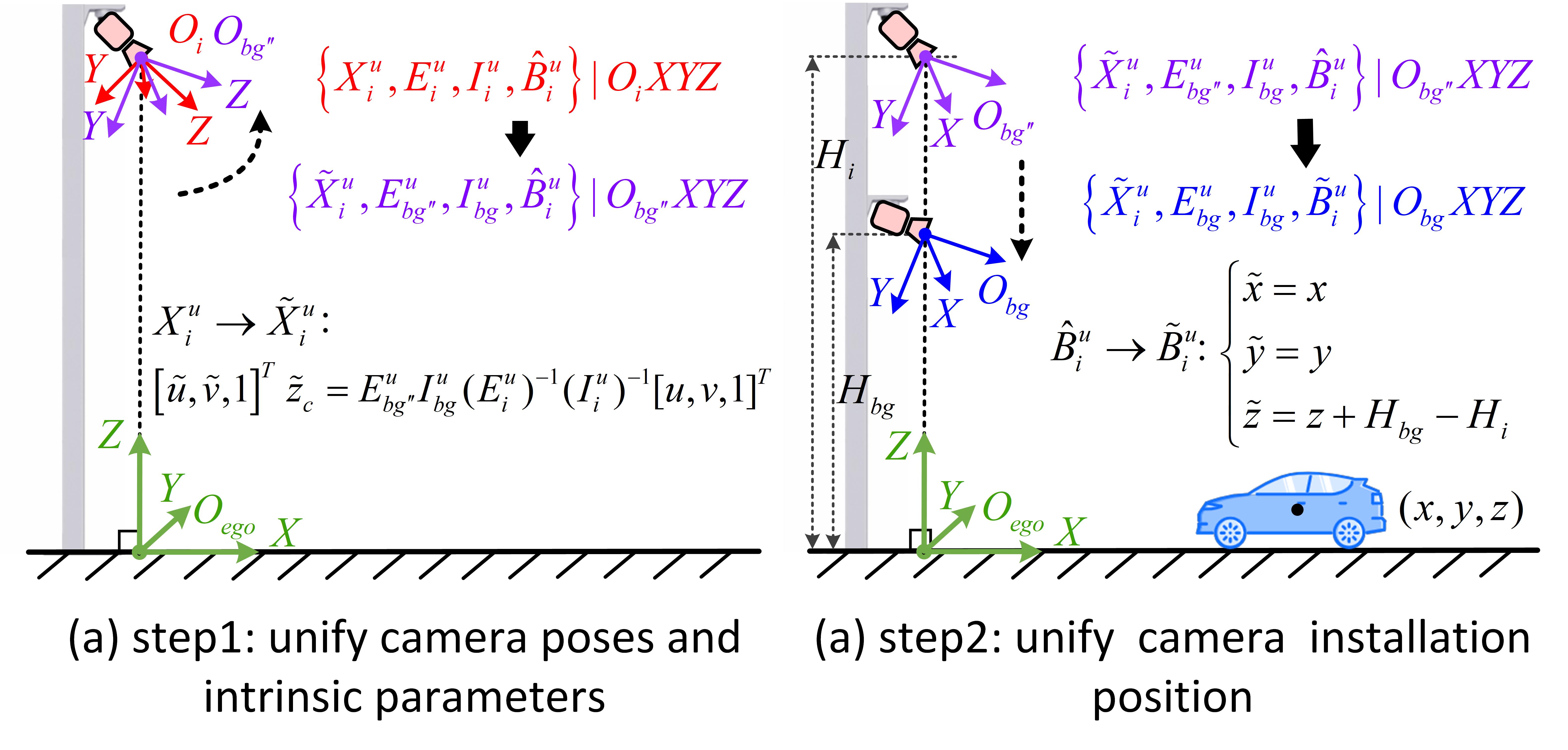}
	\caption{\textbf{The detailed steps in the rectification process.} {\color{purple}$O_{bg^{''}}XYZ$} implies the interim coordinate system with the same origin as coordinate system {\color{red}$O_iXYZ$} and same poses as coordinate system {\color{blue}$O_{bg}XYZ$}.}
\label{fig:rectify_total_steps}
\vspace{-0.15cm}
\end{figure}

\noindent \textbf{Rectify.}
Before the combination process in the Semi-supervised Data Generation Pipeline (SSDG), it is necessary to ensure consistent camera parameters between the source data $\left\{X_i^u, E_i^u, I_i^u, \hat{B}_i^u\right\}$  and background data $\left\{X_{bg}^u, E_{bg}^u, I_{bg}^u\right\}$. In the rectify process, we aim to rectify the source data from $\left\{X_i^u, E_i^u, I_i^u, \hat{B}_i^u\right\}$ to $\left\{\tilde{X}_i^u, E_{bg}^u, I_{bg}^u, \tilde{B}_i^u\right\}$, where $\tilde{X}_i^u$ is the rectified image, $\tilde{B}_i^u$ represents the rectified pseudo labels. As can be seen in Fig.~\ref{fig:problem_of_rectify} (a) and (b), the rectified data, on the one hand, maintains exactly the same camera intrinsic and extrinsic parameters as the background data, and on the other hand, the pseudo labels, when projected onto the image using camera parameters, are appropriately aligned with the corresponding objects on the image.


This process is further subdivided into two steps: (i) step 1: unify camera poses and intrinsic parameters, and (ii) step 2: unify the camera installation position. As illustrated in Fig.~\ref{fig:rectify_total_steps} (a), step 1 can be formulated as the rectification from the source data $\left\{X_i^u, E_i^u, I_i^u, \hat{B}_i^u\right\}$ to an interim data $\left\{\tilde{X}_i^u, E_{bg^{''}}^u, I_{bg}^u, \hat{B}_i^u\right\}$. $E_{bg^{''}}^u$ signifies interim extrinsic parameters of the camera. The corresponding camera coordinate system maintains the same poses as the background data and shares a coincident coordinate origin with the source data. The rectified image $\tilde{X}_i^u$ is generated as as follows:
\begin{equation}
    \left[\tilde{u},\tilde{v},1\right]^T \tilde{z} = I_{bg}^u E_{bg^{''}}^u (E_i^u)^{-1}(I_i^u)^{-1}[u,v,1]^T
\label{con:eq10}
\end{equation}
where $\left(\tilde{u}, \tilde{v}\right)$ implies pixel coordinates on the rectified image $\tilde{X}_i^u$, and $\left(u, v\right)$ is for the source image $X_i^u$.

As illustrated in Fig.~\ref{fig:rectify_total_steps} (b), step 2 can be formulated as the rectification from the interim data $\left\{\tilde{X}_i^u, E_{bg^{''}}^u, I_{bg}^u, \hat{B}_i^u\right\}$ to the rectified data $\left\{\tilde{X}_i^u, E_{bg}^u, I_{bg}^u, \tilde{B}_i^u\right\}$. In this step, we simultaneously rectify the camera extrinsic parameters from $E_{bg^{''}}^u$ to $E_{bg}^u$ and pseudo labels from $\hat{B}_i^u$ to $\tilde{B}_i^u$ as follows:
\begin{eqnarray}
    \tilde{B}_i^u&=& \left\{\tilde{b}_i^k \right\}_{k=1}^{n^r} \nonumber \\
    \tilde{b}_i^k &=& \left(x,y,z+H_{bg}-H_i,h,w,l,conf\right),
\label{con:eq11}
\end{eqnarray}
where $H_i$ and $H_{bg}$ represent the cameras' installation height from the ground of source data and background data, respectively. $n^r$ $\left(n^r \leq n^p\right)$ is the number of pseudo labels within the rectified image.

This simultaneous adjustment ensures the proper alignment of pseudo labels with their corresponding objects on the image when projected using the camera parameters. 


\noindent \textbf{SAM.}
We use the Segment Anything (SAM)~\cite{kirillov2023segment} model to help generate multi-class segmentation masks $M_{gt}^{seg}$ for the previous source data, represented as follows:
\begin{equation}
\left(M_{gt}^{seg}\right)_i=F_{SAM} \left(\tilde{X}_i^u,\tilde{B}_i^u\right) = \left\{m_i^k \right\}_{k=1}^{n^r}
\label{con:eq12}
\end{equation}
where $F_{SAM}$ is the segment anything model, using the 2D boxes projected by $\tilde{B}_i^u$ on the image as prompt. $m_i^k$ is an instance-level mask for each 2D box prompt.

The generated masks are not only used as labels for supervised training of BMS but also for selecting object foreground regions in the combination process.

\noindent \textbf{Combination.}
This process is dedicated to generating combined data $\left\{X_{comb}^u, E_{comb}^u, I_{comb}^u, B_{comb}^u, \left(M_{gt}^{seg}\right)_{comb}\right\}$ by merging a few frames of rectified source data $\left\{\tilde{X}_i^u, E_{bg}^u, I_{bg}^u, \tilde{B}_i^u, \left(M_{gt}^{seg}\right)_{i}\right\}_{i=1}^{N^s}$ and empty background data $\left\{X_{bg}^{u}, E_{bg}^{u}, I_{bg}^{u} \right\}$. The specific process can be formulated as follows:
\begin{eqnarray}
    X_{comb}^u=\sum_{i=1}^{N^s}\sum_{k=1}^{n^c}\left(m_i^k \tilde{X}_i^u + (1-m_i^k)X_{bg}^u)\right)
\label{con:eq13}
\end{eqnarray}

\begin{eqnarray}
    \left(M_{gt}^{seg}\right)_{comb}=\bigcup_{i=1}^{N^s}\bigcup_{k=1}^{n^c}m_i^k  
\label{con:eq14}
\end{eqnarray}
\begin{eqnarray}
    B_{comb}^u=\sum_{i=1}^{N^s}\sum_{k=1}^{n^c}b_i^k, E_{comb}^u=E_{bg}^u,  I_{comb}^u=I_{bg}^u,
\label{con:eq15}
\end{eqnarray}
where $N^s$ is the number of the rectified source data used in a combination process, $n^c$ $\left(n^c \leq n^r \right)$ is the number of pseudo-labels in each source data selected for the combination. Here we only consider pseudo-labels with $IoU < T_{iou}$ with other bounding boxes, $T_{iou}$ is the $IoU$ threshold.

\section{Experiments}
\subsection{Datasets}

\noindent \textbf{DAIR-V2X-I.} Yu et.al \cite{yu2022dair} introduces a large-scale dataset for vehicle-infrastructure collaborative perception collected from real-world scenarios. We focus on the DAIR-V2X-I, which only contains the images from mounted cameras to study roadside perception. It provides a homogeneous data split, with 5042 frames in the training set and 2016 frames in the validation set. To assess the algorithm's scenario generalization capacity, we introduce a heterogeneous data split for DAIR-V2X-I: 4939 frames from two scenarios for training and 2119 frames collected from a third scenario for evaluation. We follow KITTI \cite{geiger2012kitti} to use the AP$_{\text{3D}{|\text{R40}}}$ as the evaluation metric.\\
 \noindent \textbf{Rope3D.} This is another extensive roadside benchmark named Rope3D \cite{ye2022rope3d}, comprising over 50K images with 3D annotations from seventeen intersections. This dataset not only includes all the images in DAIR-V2X-I \cite{yu2022dair}, but it also provides additional data for the same scenes, including more intersections. On the one hand, we utilize the data beyond DAIR-V2X-I as unlabeled data during the training stage. On the other hand, we are also conducting experiments on this larger-scale Rope3D dataset. Here, we adhere to the proposed homologous setting, wherein 70\% of the images are allocated for training, and the remaining 30\% for testing. It is important to note that all images are randomly sampled in this process. For the heterogeneous setting, we opt to utilize 80\% of images for training purposes, reserving the unseen 20\% (collected from different intersection scenarios) for validation. As for the validation metrics, we employ the AP$_{\text{3D}{|\text{R40}}}$ and the Rope score. The Rope score serves as a comprehensive metric, encompassing AP$_{\text{3D}{|\text{R40}}}$ and other similarity metrics such as average area similarity.

\subsection{Experimental Settings}
We utilize ResNet-101\cite{he2016resnet} backbone, voxel size of [0.1m, 0.1m] in state-of-the-art comparisons. For ablation studies, ResNet-50 encoder, voxel size of [0.2m, 0.2m], and heterogeneous setting are employed. Input resolution is (864, 1536). The bird's-eye-view (BEV) grid spans [-51.2m, 51.2m] in width and [0.0m, 102.4m] in length. 
For the Semi-supervised Data Generation Pipeline (SSDG) on the DAIR-V2X-I~\cite{yu2022dair} dataset, We incorporate images from Rope3D, specifically those beyond DAIR-V2X-I, as unlabeled data during the training process. In the case of the Rope3D~\cite{ye2022rope3d} dataset, due to the lack of additional unlabeled data, we rely solely on about 30 frames of empty background data generated through \cite{8690428} from the validation set. Additionally, we use labeled data from the training set, along with annotations, as a substitute for source data.
For data augmentation, we follow~\cite{li2023bevdepth} to use random scaling and rotation in image-view. All methods undergo 35 epochs of training with the AdamW optimizer~\cite{loshchilov2017decoupled} and an initial learning rate of 2e-4.

\subsection{Comparing with state-of-the-art}

We compare the proposed SGV3D against state-of-the-art detectors, including MonoGAE \cite{yang2023monogae}, BEVFormer \cite{li2022bevformer}, BEVDepth \cite{li2023bevdepth}, and BEVHeight \cite{yang2023bevheight}, BEVHeight++ \cite{yang2023bevheight++} and CoBEV \cite{shi2023cobev}. The evaluation is conducted on DAIR-V2X-I~\cite{yu2022dair} and Rope3D~\cite{ye2022rope3d} datasets, considering heterogeneous and homologous as described above.

\noindent \textbf{Results on the homologous benchmark.} On DAIR-V2X-I~\cite{yu2022dair} dataset, Tab.~\ref{dair_sota_1} highlights the superior performance of our method compared to BEVHeight baseline across all categories. Notably, it demonstrates a substantial improvement of +6.75\% for vehicle, +5.52\% for pedestrian, and +5.03\% for cyclist. Our SGV3D ranks the first place in detecting pedestrian and cyclist and securing the second position in vehicle detection. This is attributed to the spatial settings of the bird’s-eye-view (BEV) grid. The grid of BEV perception methods in Tab.~\ref{dair_sota_1} only cover the range within 100m, which restricts predictions within the predefined range span. In this way, As in \cite{shi2023cobev}, expanding the longer range between 100-200m will lead to higher detection accuracy.

On Rope3D~\cite{ye2022rope3d} dataset, as shown in Tab.~\ref{rope3d_sota_1}, 
under easy evaluation conditions (IoU = 0.5), SGV3D secures the top position in car detection with an accuracy of 77.38\%, surpassing MonoGAE's 76.12\%. At the same time, SGV3D achieves the second-highest accuracy for big vehicle at 51.57\%, slightly trailing behind MonoGAE's 52.77\%. Even at the stricter IoU = 0.7 evaluation, SGV3D outperforms the BEVHeight baseline, showing substantial improvements of +2.88\% / +3.06\% and +2.51\% / +3.20\% in AP and $Rope_{score}$ for car and big vehicle, respectively.

\begin{table}[h]
 \scriptsize\centering\addtolength{\tabcolsep}{-4.5pt}
\caption{\textbf{Comparison with the state-of-the-art on the DAIR-V2X-I homologous benchmark.} We use \textbf{bold} to highlight the highest results and \underline{underline} for the second-highest ones. We present the results for three categories: vehicle (Veh.), pedestrian (Ped.), and cyclist (Cyc.) Average precision ($AP_{3D}$) is used as the evaluation metric. $\dagger$ implies BEV perception methods.}

\begin{tabularx}{1.0\linewidth}{l|ccc|ccc|ccc}
  \toprule
 \multirow{3}{*}{Method} &  \multicolumn{3}{c|}{$\text{Veh.}_{(IoU=0.5)}$} & \multicolumn{3}{c|}{$\text{Ped.}_{(IoU=0.25)}$} & \multicolumn{3}{c}{$\text{Cyc.}_{(IoU=0.25)}$} \\
    \cmidrule(r){2-10}
      & Easy & Mod. & Hard & Easy & Mod. & Hard & Easy & Mod. & Hard\\
\midrule
ImVoxelNet$\dagger$ \cite{rukhovich2022imvoxelnet}  &      44.78 &     37.58   &    37.55   &   6.81   &    6.75    &   6.73   &   21.06   &    13.57    &  13.17    \\
MonoGAE \cite{yang2023monogae} & \textbf{84.61}& \textbf{75.93} & \textbf{74.17} & 25.65 & 24.28 & 24.44 & 44.04 &  47.62 & 46.75 \\
CoBEV$\dagger$ \cite{shi2023cobev} &  81.20 & 68.86  & 68.99 & \underline{44.23}  &  \underline{42.31}  &  \underline{42.55} & \underline{61.28}  &  \underline{61.00}  & \underline{61.61} \\
BEVFormer$\dagger$ \cite{li2022bevformer}   &     61.37  &    50.73    &  50.73     &       16.89     &   15.82   &   15.95   &    22.16    &  22.13 &  22.06\\
BEVDepth$\dagger$ \cite{li2023bevdepth}   &     75.50  &    63.58    &   63.67    &   34.95   &   33.42     &   33.27   &   55.67   &    55.47    & 55.34 \\
BEVHeight++$\dagger$ \cite{yang2023bevheight++} &    79.31   &     68.62   &   68.68    &  42.87    &   40.88     &   41.06   &  60.76    &   60.52     & 61.01\\
BEVHeight$\dagger$  \cite{yang2023bevheight}  &   77.78    &    65.77    &     65.85  &     41.22 &   39.29     &  39.46    &   60.23   &   60.08     & 60.54\\
\midrule
 \rowcolor{cyan!30} SGV3D$\dagger$ &	
 \underline{83.44}&	\underline{72.52}&	\underline{72.81}&	\textbf{46.12}&	\textbf{44.81}&	\textbf{44.92}	&\textbf{65.84}&	\textbf{65.11}& \textbf{65.04}	\\
\rowcolor{cyan!30} w.r.t BEVHeight  &   \tg{+5.66}  &  \tg{+6.75}    &  \tg{+6.96}  &   \tg{+4.90} &  \tg{+5.52}   &  \tg{+5.46}  &   \tg{+5.61}   &  \tg{+5.03}  &  \tg{+4.50} \\
\bottomrule
\multicolumn{10}{l}{\scriptsize{-Note: All methods rely solely on images for both the training and testing phases.}}
\end{tabularx}
\vspace{-0.2cm}
  \label{dair_sota_1}
\end{table}
\begin{table}[h!t]
\scriptsize\centering\addtolength{\tabcolsep}{-3.5pt}
\caption{\textbf{Comparison with state-of-the-art methods on the Rope3D validation set in homologous settings.}Here, $Rope$ denotes $Rope_{score}$ introduced in \cite{ye2022rope3d}. $+G$ represents adopting the ground plane. We use \textbf{bold} to highlight the highest results and \underline{underline} for the second-highest ones. $\dagger$ denotes BEV perception methods.}
 \begin{tabularx}{1.\linewidth}{ l |cc|cc|cc|cc }
\toprule
\multirow{4}{*}{Method}   & \multicolumn{4}{c|}{IoU = 0.5} & \multicolumn{4}{c}{IoU = 0.7} \\ 
\cmidrule(r){2-9}
  & \multicolumn{2}{c|}{Car} & \multicolumn{2}{c|}{Big Vehicle} & \multicolumn{2}{c|}{Car} & \multicolumn{2}{c}{Big Vehicle} \\ 
\cmidrule(r){2-9}
&AP & Rope &
AP & Rope &
AP & Rope &
AP & Rope \\

\midrule
M3D-RPN(+G) \cite{he2021aug3d}    & 54.19 & 62.65 & 33.05 & 44.94 & 16.75 & 32.90 & 6.86 & 24.19  \\     
MonoDLE(+G) \cite{ma2021delving}    & 51.70 & 60.36 & 40.34 & 50.07 & 13.58 & 29.46 & 9.63 & 25.80 \\ 
Kinematic3D(+G) \cite{brazil2020kinematic}  & 50.57 & 58.86 & 37.60 & 48.08 & 17.74 &  32.90 & 6.10 & 22.88  \\ 
BEVDepth$\dagger$ \cite{li2023bevdepth}   & 69.63 & 74.70 & 45.02 & 54.64 & 42.56 & 53.05 & 21.47 & 35.82 \\
BEVFormer$\dagger$ \cite{li2022bevformer} & 50.62 & 58.78 & 34.58 & 45.16 & 24.64 & 38.71 & 10.05 & 25.56  \\
BEVHeight++$\dagger$ \cite{yang2023bevheight++} & \underline{76.12} & \underline{80.91} & \underline{50.11} & \underline{59.92} & \underline{47.03} & \underline{57.77} & \underline{24.43} & \underline{39.57} \\
BEVHeight$\dagger$ \cite{yang2023bevheight}   & 74.60 & 78.72 & 48.93 & 57.70 & 45.73 & 55.62 & 23.07 & 37.04 \\
\midrule
 \rowcolor{cyan!30} SGV3D$\dagger$ & \textbf{77.38}& \textbf{82.35}& \textbf{51.57}& \textbf{60.01}& \textbf{48.61}& \textbf{58.68}& \textbf{25.58}& \textbf{40.24} \\		
 \rowcolor{cyan!30} w.r.t BEVHeight$\dagger$  &   {\scriptsize \tg{+2.78}}  &  {\scriptsize \tg{+3.63}}    &  {\scriptsize \tg{+2.64}}  &   {\scriptsize \tg{+2.31}} &  {\scriptsize \tg{+2.88}}   &  {\scriptsize \tg{+3.06}}  &   {\scriptsize \tg{+2.51}}   &  {\scriptsize \tg{+3.20}} \\
\bottomrule
\multicolumn{9}{l}{\footnotesize{AP and Rope denote AP$_{\text{3D}{|\text{R40}}}$ and Rope$_\text{score}$ respectively.}}
\end{tabularx}
\vspace{-0.2cm}
\label{rope3d_sota_1}
\end{table}

\noindent \textbf{Results on the heterogeneous benchmark.} On DAIR-V2X-I~\cite{yu2022dair} setting, as presented in Tab.~\ref{dair_sota_2}, the performance of existing methods exhibits a substantial decline when evaluated on new scenes. Taking Vehicle (IoU=0.5) as an example, BEVFormer\cite{li2022bevformer} experiences a decrease from 50.73\% to 5.42\%, BEVHeight \cite{yang2023bevheight} similarly declines from 65.77\% to 7.86\%, and BEVHeight++ \cite{yang2023bevheight++} experiences a decrease from 68.62\% to 8.18\%. In contrast, our SGV3D maintains a performance of 50.75\%, only slightly reducing from the original 69.78\%. Remarkably, this is a significant improvement over the other leading method, surpassing by a large margin of 42.57\%, 5.87\%, and 14.89\% for vehicle, pedestrian, and cyclist, respectively. These results underscore the remarkable scenario generalization capability of our approach.

For the Rope3D~\cite{ye2022rope3d} dataset, as shown in Tab.~\ref{rope3d_sota_2}, we conduct a comparative evaluation of SGV3D against other leading vision-based 3D detectors on the
Rope3D heterologous benchmark. Our method achieves the best detection accuracy of (10.87\% vs. 6.59\%) for car and (5.82\% vs. 2.78\%) for big vehicle under the strict evaluation conditions characterized by IoU = 0.7. Under the easier IoU = 0.5, SGV3D surpasses the BEVHeight baseline with improvements of +14.48\% / +15.44\% and +12.41\% / +12.57\% in AP / $Rope_{score}$ for car and big vehicle, respectively.

\begin{table}[h]
 \scriptsize\centering\addtolength{\tabcolsep}{-5.5pt}
\caption{\textbf{Comparison with the state-of-the-art on the DAIR-V2X-I heterologous benckmark.} We present the results for vehicle (Veh.), pedestrian (Ped.), and cyclist (Cyc.) Average precision ($AP_{3D}$) is used as the evaluation metric. We use \textbf{bold} to highlight the highest results and \underline{underline} for the second-highest ones. $\dagger$ implies BEV perception methods.}

 \begin{tabularx}{1.0\linewidth}{l|ccc|ccc|ccc}
  \toprule
 \multirow{3}{*}{Method} &  \multicolumn{3}{c|}{$\text{Veh.}_{(IoU=0.5)}$} & \multicolumn{3}{c|}{$\text{Ped.}_{(IoU=0.25)}$} & \multicolumn{3}{c}{$\text{Cyc.}_{(IoU=0.25)}$} \\
    \cmidrule(r){2-10}
      & Easy & Mod. & Hard & Easy & Mod. & Hard & Easy & Mod. & Hard \\
\midrule
MonoGAE \cite{yang2023monogae}    &  \underline{10.50} &   \underline{9.15}&    \underline{8.97}&    2.04 &       1.57 &   1.64  &     4.32&     3.73& 3.79     \\
BEVFormer$\dagger$ \cite{li2022bevformer}   & 5.92  & 5.42  & 5.39  & 1.82 & 1.60  & 1.66  & 3.59 & 3.42   & 3.48 \\
BEVDepth$\dagger$ \cite{li2023bevdepth}    &      2.82&     2.48&     2.37 &    0.84&      0.73&     0.72&     1.53&       1.37& 1.34      \\
BEVHeight++$\dagger$ \cite{yang2023bevheight++} &     9.05 &   8.18 &   8.16  &    \underline{2.86}&   \underline{2.45}&    \underline{2.51}&    \underline{5.23}&    \underline{4.89}& \underline{4.92}    \\
BEVHeight$\dagger$ \cite{yang2023bevheight}   & 8.30&	7.86&	7.87&	2.74&	2.34&	2.31& 5.07&	4.81 &4.87\\
\midrule
 \rowcolor{cyan!30} SGV3D$\dagger$ &	
 \textbf{62.59}&	\textbf{50.75}&	\textbf{50.68}&	\textbf{8.99}&	\textbf{8.32}&	\textbf{8.53}	&\textbf{19.99}&	\textbf{19.78}& \textbf{19.82} \\
\rowcolor{cyan!30} w.r.t BEVHeight$\dagger$  &   \tg{+54.29}  &  \tg{+42.89}    &  \tg{+42.71}  &   \tg{+6.25} &  \tg{+5.98}   &  \tg{+6.22}  &   \tg{+14.92}   &  \tg{+14.97}  &  \tg{+14.95} \\

\bottomrule
\end{tabularx}
\vspace{-0.2cm}
\label{dair_sota_2}
\end{table}
\begin{table}[h]
\scriptsize\centering\addtolength{\tabcolsep}{-4.3pt}
\caption{\textbf{Comparison with state-of-the-art methods on the Rope3D validation set in heterologous settings.} Here, $Rope$ denotes $Rope_{score}$ introduced in \cite{ye2022rope3d}. $+G$ represents adopting the ground plane. We use \textbf{bold} to highlight the highest results and \underline{underline} for the second-highest ones. $\dagger$ represents BEV perception methods.}
 \begin{tabularx}{1.\linewidth}{ l |cc|cc|cc|cc }
\toprule
\multirow{4}{*}{Method}   & \multicolumn{4}{c|}{IoU = 0.5} & \multicolumn{4}{c}{IoU = 0.7} \\ 
\cmidrule(r){2-9}
  & \multicolumn{2}{c|}{Car} & \multicolumn{2}{c|}{Big Vehicle} & \multicolumn{2}{c|}{Car} & \multicolumn{2}{c}{Big Vehicle} \\ 
\cmidrule(r){2-9}
&AP & Rope &
AP & Rope &
AP & Rope &
AP & Rope \\

\midrule
M3D-RPN(+G) \cite{he2021aug3d}     & 21.75 & 36.4  & 21.49 & 35.49 & 6.05 & 23.84 & \underline{2.78}  & \underline{20.82}  \\
MonoDLE(+G) \cite{ma2021delving}     & 19.08 & 33.72 & \underline{19.76} & \underline{33.07} & 3.77 & 21.42 & 2.31 & 19.55  \\ 
Kinematic3D(+G) \cite{brazil2020kinematic} & 23.56 & 37.05 & 13.85 & 28.58 & 5.82 & 23.06 & 1.27 & 18.92  \\ 
BEVDepth$\dagger$ \cite{li2023bevdepth}    & 9.00  & 25.80 & 3.59  & 20.39 & 0.85 & 19.38 & 0.30 & 17.84 \\ 
BEVFormer$\dagger$ \cite{li2022bevformer}   & 25.98 & 39.51 & 8.81  & 24.67 & 3.87 & 21.84 & 0.84 & 18.42 \\
CoBEV$\dagger$ \cite{shi2023cobev} &  \underline{31.25}&  \underline{43.74}&  16.11&  30.73&  \underline{6.59}&  \underline{24.01}&  2.26&  19.71\\
BEVHeight$\dagger$ \cite{yang2023bevheight}   & 29.65&  42.48&  13.13&  28.08&  5.41&  23.09&  1.16&  18.53  \\
\midrule
 \rowcolor{cyan!30} SGV3D$\dagger$ & \textbf{44.13}& \textbf{57.92}& \textbf{25.54}& \textbf{40.65}& \textbf{10.87}& \textbf{29.23}& \textbf{5.82}& \textbf{23.45} \\
 \rowcolor{cyan!30} w.r.t BEVHeight$\dagger$  &   {\scriptsize \tg{+14.48}}  &  {\scriptsize \tg{+15.44}}    &  {\scriptsize \tg{+12.41}}  &   {\scriptsize \tg{+12.57}} &  {\scriptsize \tg{+5.46}}   &  {\scriptsize \tg{+6.14}}  &   {\scriptsize \tg{+4.66}}   &  {\scriptsize \tg{+4.92}} \\
\bottomrule
\multicolumn{9}{l}{\footnotesize{AP and Rope denote AP$_{\text{3D}{|\text{R40}}}$ and Rope$_\text{score}$ respectively.}}
\end{tabularx}
\vspace{-0.2cm}
\label{rope3d_sota_2}
\end{table}

\begin{figure*}[t!]
\centering
\setlength{\abovecaptionskip}{0.0cm}
\includegraphics[width=17.4cm]{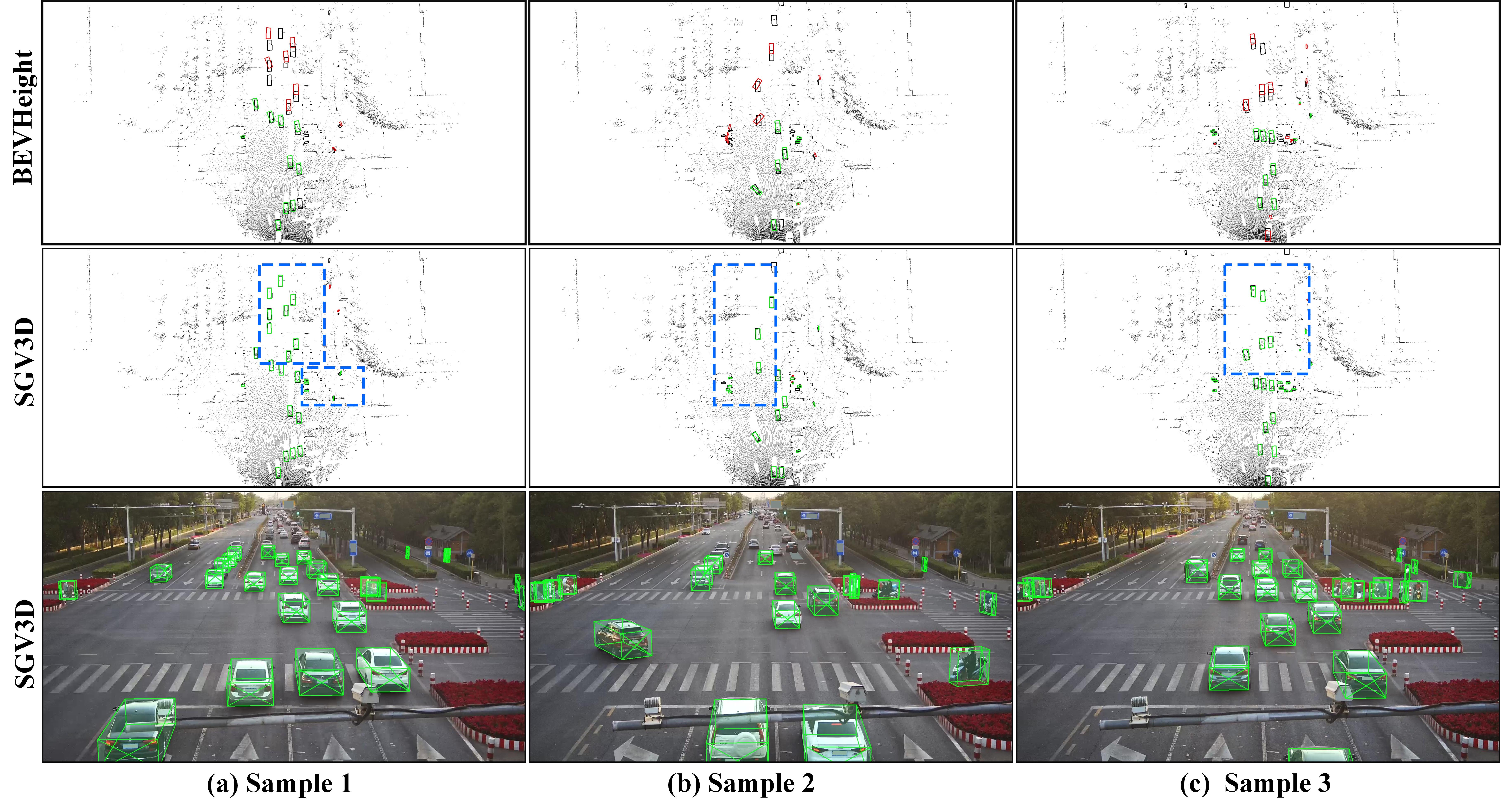}
	\caption{\textbf{Visualization Results of BEVHeight and our proposed SGV3D on the DAIR-V2X-I heterologous benckmark.} We use boxes in \textbf{{\color{red}red}} to represent false positives,  \textbf{{\color{green}green}} boxes for truth positives, and \textbf{{\color{black}black}} for the ground truth. 
    Point cloud is only used for visualization. Samples (a-c) are sourced from the new scenario in the DAIR-V2X-I validation set. 
    The \textbf{{\color{blue}blue}} rectangles remark the objects where our algorithm can significantly outperform previous algorithms.
}
\label{fig:visual}
\end{figure*}

\begin{figure*}[t!]
\centering
\setlength{\abovecaptionskip}{0.0cm}
\includegraphics[width=17.4cm]{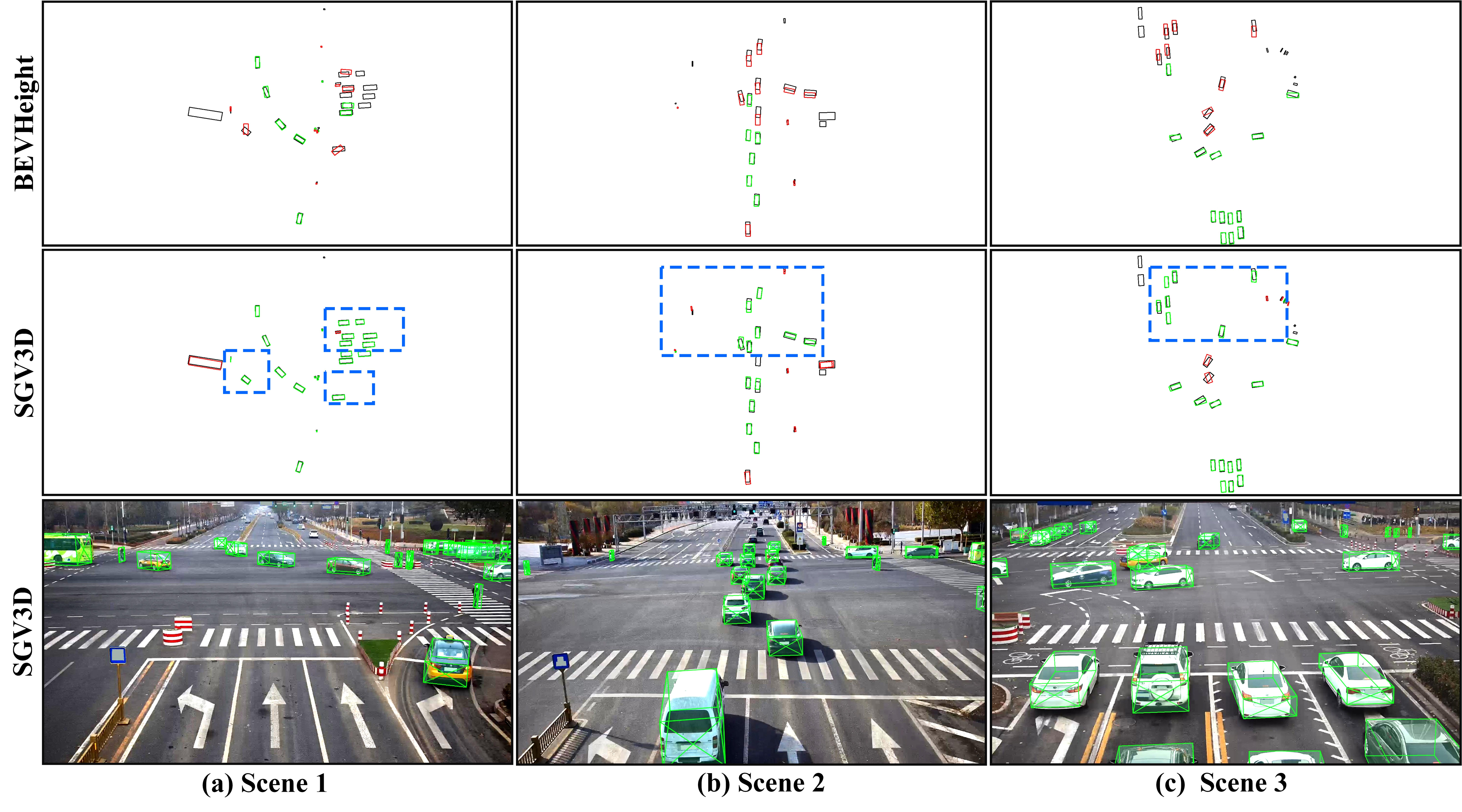}
\caption{\textbf{Visualization Results of BEVHeight and our proposed SGV3D on the
Rope3D heterologous benckmark.} We utilize \textbf{{\color{red}red}} boxes to denote false positives, \textbf{{\color{green}green}} boxes for true positives, and \textbf{{\color{black}black}} for ground truth. (a)-(c) represents three different newly roadside scenarios. We use
\textbf{{\color{blue}Blue}} rectangles to highlight instances where our SGV3D significantly outperforms previous BEVHeight approach.
} 
\label{fig:visual_rope3d}
\end{figure*}

\noindent \textbf{Visualization Results.} For DAIR-V2X-I~\cite{yu2022dair} dataset, as shown in Fig.~\ref{fig:visual}, we present the results of BEVHeight baseline~\cite{yang2023bevheight} and our new proposed SGV3D in the image view and BEV space under heterogeneous settings. All these samples are sourced from the scenario in the DAIR-V2X-I validation set. From the samples in (a-c), the predictions of BEVHeight exhibit significant deviations and omissions in relation to ground truth for middle and distant objects. In contrast, the results of our method perform much better, keeping the correct position with the ground truth. As for the results in (a), We can further observe that SGV3D exhibits fewer missed detections for small objects such as pedestrian and cyclist.

For Rope3D~\cite{ye2022rope3d} dataset, as depicted in Fig.~\ref{fig:visual_rope3d}, we present the results of BEVHeight \cite{yang2023bevheight} and our Approach in the image view and BEV space, respectively. All these scenes are sourced from the Rope3D validation set in heterogeneous settings. As observed in (a-c), there tends to be a significant offset with the ground truth over middle and long distances in BEVHeight \cite{yang2023bevheight} detections. In contrast, our method produces more precise predictions. For example, the blue dashed rectangles highlight that our SGV3D demonstrates a substantial improvement in detecting distant, occluded vehicles.

\subsection{Ablation Study}
In this section, we perform ablation studies to investigate the effects of each component of our framework. All the ablation experiments are conducted on the DAIR-V2X-I heterogeneous benchmark.

\noindent \textbf{BMS and SSDG modules.} 
We conduct ablation experiments on the effects of the background-suppressed module (BMS) and semi-supervised data generation pipeline (SSDG) modules. As shown in Tab.~\ref{module_ablation}, 
The baseline model, devoid of the BMS and SSDG modules, yields detection accuracy of 6.04\%, 1.58\%, and 3.59\% for vehicle, pedestrian, and cyclist, respectively. Several key observations arise: (i) The integration of the BMS module results in performance improvements of +8.53\%, +2.46\%, and +3.33\% for the corresponding classes, without the use of any unlabeled data. (ii) Additionally, the inclusion of the SSDG module further elevates the average precision up to 49.03\%, 6.92\%, and 17.29\%, underscoring the efficacy of our designs.
\begin{table}[h]
 \scriptsize\centering\addtolength{\tabcolsep}{-2.8pt}
\caption{\textbf{Ablation studies on the effects of BMS and SSDG modules in SGV3D on the DAIR-V2X-I heterologous setting.}} 
 \begin{tabularx}{1.0\linewidth}{c|c|ccc|ccc|ccc}
  \toprule
 \multirow{3}{*}{BSM} &  \multirow{3}{*}{SSDG} & \multicolumn{3}{c|}{$\text{Veh.}_{(IoU=0.5)}$} & \multicolumn{3}{c|}{$\text{Ped.}_{(IoU=0.25)}$} & \multicolumn{3}{c}{$\text{Cyc.}_{(IoU=0.25)}$} \\
    \cmidrule(r){3-11}
           &  & Easy & Mod. & Hard & Easy & Mod. & Hard & Easy & Mod. & Hard \\
\midrule
 ~& ~& 6.21&	6.04&	6.05&	1.76& 1.58& 1.63& 3.89&	3.59&	3.61 \\
\checkmark &  & 20.08&     14.57&     14.55&     3.51&     4.04&      4.17&      7.37&      6.92&        6.93      \\
\checkmark & \checkmark &  \textbf{59.38}&  \textbf{49.03}&  \textbf{46.96}&  \textbf{7.25}&  \textbf{6.92}&  \textbf{7.12}&  \textbf{17.96}&  \textbf{17.29}&  \textbf{17.23}  \\ 
\bottomrule
  \end{tabularx}
  \label{module_ablation}
\end{table}

\noindent \textbf{The hyper-parameters in BMS module.} 
We explored the impact of varying the hyperparameter $T_{fg}$ within the Background-Suppressed Module (BMS). As depicted in Tab. \ref{bg_conf_ablation}, the outcomes unequivocally demonstrate that the optimal prediction performance is attained when employing a confidence threshold of 0.55.
\begin{table}[h]
\footnotesize
  \centering\addtolength{\tabcolsep}{-2.7pt}
\caption{\textbf{Ablation studies on the effects of hyper-parameter $T_{fg}$ in the Background-Suppressed Module (BMS).} Here, $T_{fg}$ denotes the threshold of background. 
}
 \begin{tabularx}{1.\linewidth}{ c|ccc|ccc|ccc}
\toprule
\multirow{2}{*}{$T_{fg}$} & \multicolumn{3}{c|}{$\text{Veh.}_{(IoU=0.5)}$} & \multicolumn{3}{c|}{$\text{Ped.}_{(IoU=0.25)}$} & \multicolumn{3}{c}{$\text{Cyc.}_{(IoU=0.25)}$}\\ 
\cmidrule(r){2-10}
~& Easy & Mod. & Hard & Easy & Mod. & Hard & Easy & Mod. & Hard\\
\midrule
0.70 &  56.58&  48.87&  45.88&  6.92&  6.54&  6.73&  17.84&  17.12 & 17.04  \\
0.55 &  \textbf{59.38}&  \textbf{49.03}&  \textbf{46.96}&  \textbf{7.25}&  \textbf{6.92}&  \textbf{7.12}&  \textbf{17.96}&  \textbf{17.29} &  \textbf{17.23}  \\ 
0.40 &  55.69&  47.92&  45.91&  5.64&  5.77&  6.02&  16.92&  16.84 &  16.81\\ 
0.25 &  53.81&  45.14&  45.09&  5.94&  5.54&  5.85&  16.05&  15.71&   15.63  \\ 			
\bottomrule
\end{tabularx}
\label{bg_conf_ablation}
\end{table}

\begin{table}[h]
\footnotesize
  \centering\addtolength{\tabcolsep}{-4.2pt}
\caption{\textbf{Ablation studies on the effects of hyper-parameters in Semi-supervised Data Generation Pipeline (SSDG).} Here, $T_{conf}$ implies the confidence threshold to select high-quality pseudo labels during the Pseudo-Labeling process. $T_{iou}$ is the IoU threshold, determining whether to apply the paste operation for the instance foreground during the Combination stage}
 \begin{tabularx}{1.\linewidth}{ c|c|ccc|ccc|ccc}
\toprule
\multirow{2}{*}{$T_{conf}$} & \multirow{2}{*}{$T_{iou}$}  & \multicolumn{3}{c|}{$\text{Veh.}_{(IoU=0.5)}$} & \multicolumn{3}{c|}{$\text{Ped.}_{(IoU=0.25)}$} & \multicolumn{3}{c}{$\text{Cyc.}_{(IoU=0.25)}$}\\ 
\cmidrule(r){3-11}
& & Easy & Mod. & Hard & Easy & Mod. & Hard & Easy & Mod. & Hard\\
\midrule
0.70  & 0.05 &  58.45& 48.24&  46.35&  7.19&  6.84&  7.03&  16.04&  16.50 & 16.40   \\
0.70  & 0.25 &  \textbf{59.38}&  \textbf{49.03}&  \textbf{46.96}&  \textbf{7.25}&  \textbf{6.92}&  \textbf{7.12}&  \textbf{17.96}&  \textbf{17.29}&  \textbf{17.23} \\ 
0.70  & 0.45 &  57.72&  47.57&  46.19&  7.19&  6.86&  7.09&  17.02&  16.32  & 15.38  \\ 
0.60  & 0.25 &  54.54&  45.83&  43.83&  5.93&  5.19&  5.13&  16.21& 15.85 & 15.81   \\
0.80  & 0.25 &  56.30&  45.82&  43.84&  6.01&  5.97&  6.16&  16.56&  16.19 & 16.20 \\ 
			
\bottomrule
\end{tabularx}
\vspace{-0.1cm}
\label{conf_ablation}
\end{table}

\begin{table}[t]
\footnotesize
  \centering\addtolength{\tabcolsep}{-3.0pt}
\caption{\textbf{Ablation studies on the effect of the number of rounds in the multi-round training paradigm.}
}
 \begin{tabularx}{1.\linewidth}{ c|ccc|ccc|ccc}
\toprule
\multirow{2}{*}{Round}  & \multicolumn{3}{c|}{$\text{Veh.}_{(IoU=0.5)}$} & \multicolumn{3}{c|}{$\text{Ped.}_{(IoU=0.25)}$} & \multicolumn{3}{c}{$\text{Cyc.}_{(IoU=0.25)}$}\\ 
\cmidrule(r){2-10}
~ &Easy & Mod. & Hard & Easy & Mod. & Hard & Easy & Mod. & Hard\\
\midrule
1 & 53.58&  42.37&  42.33&  5.55&  5.38&  5.55&  12.52&  12.48&  11.51   \\
2 & 54.62&  44.92&  42.95&  6.51&  6.02&  6.19&  15.69&  15.39&  14.44   \\
3 & 57.28&  46.44&  46.36&  6.67&  6.18&  5.85&   16.04&  16.51&  16.39   \\
4 & 56.76&  48.52&  46.51&  6.75&  6.58&  6.73&  17.57&  17.01&  16.92   \\
5 & \textbf{59.38}&  \textbf{49.03}&  \textbf{46.96}&  \textbf{7.25}&  \textbf{6.92}&  \textbf{7.12}&  \textbf{17.96}&  \textbf{17.29}&  \textbf{17.23} \\ 
\bottomrule
\end{tabularx}
\vspace{-0.1cm}
\label{round_ablation}
\end{table}

\begin{table}[h!t]
\footnotesize
  \centering\addtolength{\tabcolsep}{-4.7pt}
\caption{\textbf{Ablation studies on the scales of unlabeled data.}  
}
 \begin{tabularx}{1.\linewidth}{c|ccc|ccc|ccc}
\toprule
\multirow{2}{*}{Unlabeled Data }  & \multicolumn{3}{c|}{$\text{Veh.}_{(IoU=0.5)}$} & \multicolumn{3}{c|}{$\text{Ped.}_{(IoU=0.25)}$} & \multicolumn{3}{c}{$\text{Cyc.}_{(IoU=0.25)}$}\\ 
\cmidrule(r){2-10}
~ &Easy & Mod. & Hard & Easy & Mod. & Hard & Easy & Mod. & Hard\\

\midrule
25\% & 39.49&  34.36&  34.29&  5.84&  5.58& 5.64 &  12.47& 12.23 & 12.14  \\
50\% & 47.85&  41.51&  39.67&  5.66&  6.10&  6.32&  13.82 & 13.60 & 13.51    \\
75\% & 55.23&  46.69&  43.63&  6.68&  6.39&  6.51& 15.39 &  15.34&  15.32   \\
100\% & \textbf{59.38}&  \textbf{49.03}&  \textbf{46.96}&  \textbf{7.25}&  \textbf{6.92}&  \textbf{7.12}&  \textbf{17.96}&  \textbf{17.29}&  \textbf{17.23}  \\ 	
\bottomrule
\end{tabularx}
\vspace{-0.3cm}
\label{inject_ablation}
\end{table}

\noindent \textbf{The hyper-parameters in SSDG module.} We examine the impacts of various hyperparameters in the semi-supervised data generation pipeline (SSDG) module, specifically $T_{conf}$ for distinguishing high-quality pseudo labels and $T_{iou}$ for determining instance candidates during the combination process. As indicated in Table~\ref{conf_ablation}, optimal performance is realized when $T_{conf}$ is set to 0.7, and $T_{iou}$ is set to 0.25.

\noindent \textbf{Round number in the multi-round training paradigm.} 
In adherence to the multi-round training scheme, the determination of the number of semi-supervised training rounds stands out as a crucial hyper-parameter. We conduct ablation studies on this parameter, as illustrated in Fig. \ref{round_ablation}. The results reveal substantial accuracy improvements in the initial four training rounds, while the average precision exhibits a gradual and slower growth in the subsequent fifth round.

\noindent \textbf{The scale of unlabeled data.} The effects of the scale of unlabeled data used in the Semi-Supervised Data Generation (SSDG) module are presented in Tab. \ref{inject_ablation}, where 25\%, 50\%, 75\%, and 100\% indicate that 9.5K, 19K, 28.5K, and 38K unlabeled image data are utilized in the SSDG pipeline. In comparison with the results obtained using 9.5K unlabeled data, utilizing the entire unlabeled dataset further enhances accuracy from 34.36\% to 49.03\%, 5.38\% to 6.92\%, and 12.23\% to 17.29\% for car, pedestrian, and cyclist, respectively. This suggests that incorporating more unlabeled data is effective in improving performance.



\section{Conclusion}
In this paper, a novel scenario generalization framework for vision-based roadside 3D Object detection, named SGV3D is proposed to achieve great transferability from labeled scenes to new scenes. It comprises a background-suppressed module (BMS) to reduce algorithm overfitting to the backgrounds of labeled scenes and a semi-supervised data generation pipeline (SSDG) to generate diverse, well-labeled images under different camera poses for the training stage. A comprehensive evaluation is performed on the DAIR-V2X-I dataset using the non-overlapping data from Rope3D as unlabeled data. SGV3D achieves state-of-the-art results, especially for pedestrian and cyclist in homologous settings. Moreover, it significantly surpasses other leading detectors by a large margin of +42.57\%, +5.87\%, and +14.89\% for three categories in heterologous settings.  Under the larger-scale Rope3D heterogeneous benchmark, we further achieve significant performance improvements. We hope our work can shed light on the study of roadside perception with great scenario generalization ability.

{
    \small
    \bibliographystyle{ieeenat_fullname}
    \bibliography{main}
}


\end{document}